\documentclass[10pt,twocolumn,letterpaper]{article}

\usepackage{iccv}
\usepackage{times}
\usepackage{epsfig}
\usepackage{graphicx}
\usepackage{amsmath}
\usepackage{amssymb}

\usepackage[pagebackref=true,breaklinks=true,letterpaper=true,colorlinks,bookmarks=false]{hyperref}

\iccvfinalcopy 

\def\iccvPaperID{****} 

\ificcvfinal\fi
\begin{document}
	
	\title{Assisting human experts in the interpretation of their visual process: A case study on assessing copper surface adhesive potency}%
	
	\author{
		Tristan Hascoet\\
		Kobe University\\
		{\tt\small tristan@people.kobe-u.ac.jp}
		\and
		Xuejiao Deng\\
		Kobe University\\
		{\tt\small dengxuejiao1005@yahoo.co.jp}
		\and
		Kiyoto Tai\\
		MEC Co., Ltd.\\
		{\tt\small tai295@mec-np.com}
		\and
		Yuji Adachi \\
		MEC Co., Ltd. \\
		\and
		Sachiko Nakamura\\
		MEC Co., Ltd.\\
		\and
		Tomoko Hayashi\\
		MEC Co., Ltd.\\
		\and
		Mari Sugiyama\\
		MEC Co., Ltd.\\
		\and
		Yasuo Ariki\\
		Kobe University\\
		\and
		Tetusya Takiguchi\\
		Kobe University\\
	}
	
	\maketitle

\begin{abstract}
Deep Neural Networks are often though to lack interpretability due to the distributed nature of their internal representations. 
In contrast, humans can generally justify, in natural language, for their answer to a visual question with simple common sense reasoning. 
However, human introspection abilities have their own limits as one often struggles to justify for the recognition process behind our lowest level feature recognition ability: 
for instance, it is difficult to precisely explain why a given texture seems more characteristic of the surface of a finger nail rather than a plastic bottle.
In this paper, we showcase an application in which deep learning models can actually  help human experts justify for their own low-level visual recognition process: 
We study the problem of assessing the adhesive potency of copper sheets from microscopic pictures of their surface. 
Although highly trained material experts are able to qualitatively assess the surface adhesive potency, 
they are often unable to precisely justify for their decision process. 
We present a model that, under careful design considerations, 
is able to provide visual clues for human experts to understand and justify for their own recognition process. 
Not only can our model assist human experts in their interpretation of the surface characteristics, 
we show how this model can be used to test different hypothesis of the copper surface response to different manufacturing processes. 
\end{abstract}
	
\section{Introduction}

Humans are experts in communicating the reasoning process behind their answer to visual questions.
For instance, on typical Visual Question Answering (VQA) samples\cite{antol2015vqa,zhang2016yin,goyal2017making}, 
human annotators are often able to convincingly justify, in natural language, the reason 
behind their answer to a certain visual question using simple common sense reasoning.
In contrast, deep Learning models are often viewed as black box predictors lacking interpretability 
in the sense that existing tools often fail to explain the decision process behind the model’s predictions.
For instance, a deep learning model trained end-to-end on a VQA dataset may be able to provide the same answer as its
human counterpart, but the process through which the model reaches this answer is entirely opaque.

While it is true that humans can justify for their answers on high level reasoning tasks, 
humans also often fail to explain the process behind their low-level feature recognition ability:
For example, precisely defining the nature of a specific texture 
(what are the defining features of a plastic or a wooden surface?) 
or specific low-level part attributes exhibiting large intra-class variations 
(what is the defining features of a ``leg'' or a ``wing''?) is a very difficult task.
Humans constantly perform such low-level visual recognition tasks 
while being unable to precisely justify for their own recognition process.
	
In this paper, we present one very practical instance of such a situation in the Printed Circuit Boards (PCB) industry, 
in which expert material scientists are tasked with assessing the adhesive potency of copper surfaces.
We propose a model that, under careful design considerations, is able to provide visual clues 
for human experts to understand and justify for their decision process.
	
The proposed model is designed so that a subset of its internal representations carry semantically meaningful 
information that can be visualized and easily interpreted by humans.
Providing these visual clues, however, comes with the cost of imposing additional constraints on the architecture,
which we found to degrade the model accuracy:
Indeed, we found that networks with unrestricted architectures, 
(which do not provide interpretable features)
perform better than network architectures restricted so as to provide 
semantically meaningful representations.
This is because, as we restrict the architecture of the model, 
we formulate an assumption on the impact of the manufacturing process 
on the surface statistics which may not hold in reality.
This result suggests an inherent trade-off between  
expressivity and the explainability in designing model architecture.
	
While the degradation of the model accuracy is problematic from a performance perspective,
it offers an interesting opportunity from an explainability perspective:
As the model accuracy degrades due to the inadequacy of the assumption 
made by the model,
we can use the model accuracy as a proxy metric 
for the adequacy of different assumptions.
This allows us to quantitatively assess different assumptions regarding the impact 
of manufacturing processes on the copper surface. 
This may prove useful to quantify the impact of manufacturing process on cooper surface adhesive potency
and eventually help optimize the manufacturing process.
	
In essence, the argument this paper is aiming for is as follows: 
although deep learning models lack the ``common sense reasoning'' abilities of humans,
and the powerful formalism of natural language to communicate and justify for their decision process, 
they can provide useful tools to visualize and explain low-level recognition processes.

In practice, the contribution of this paper is as follows:
\begin{enumerate}
\item  We formalize a segmentation procedure based on a probabilistic weak label segmentation framework.
\item  We introduce a formalism to show how the model accuracy can be used as a proxy metric to quantify the validity of different assumptions on the dataset.
\end{enumerate}
	
The remainder of this paper is organized as follows:
In Section 2, we present some background information on the motivation for this project:
We start by discussing the importance of copper surface adhesive potency,
and detail the dataset used in our experiments. 
Section 3 details our contribution.
Section 4 briefly relates our work to different research topics 
and Section 5 presents the results of our experiments.
Finally, Section 6 further discusses the relevance of our results, insisting on the limitations of our assumptions to conclude this paper.
	
\section{Background}

Printed circuit boards (PCBs) are an integral part of a wide variety of electronic devices, 
including industrial and household appliances (e.g. TV and PC), 
mobile communication devices and automobiles. 
PCBs play an important role in electrical connection between electronic components. 
Copper has been used in the PCBs industry as the conducting material, 
and the electric copper circuits are isolated from each other by insulators (solder resist, prepreg etc). 
A multilayered PCBs have a laminate having a plurality of electro-conductive 
layers with insulating layers interposed therebetween. 
So there are many interfaces related to the copper and resins in PCBs. 
Figure 1 illustrates the organization of such an electric circuit.

\begin{figure}[h]
\centering
\includegraphics[width=0.9\linewidth]{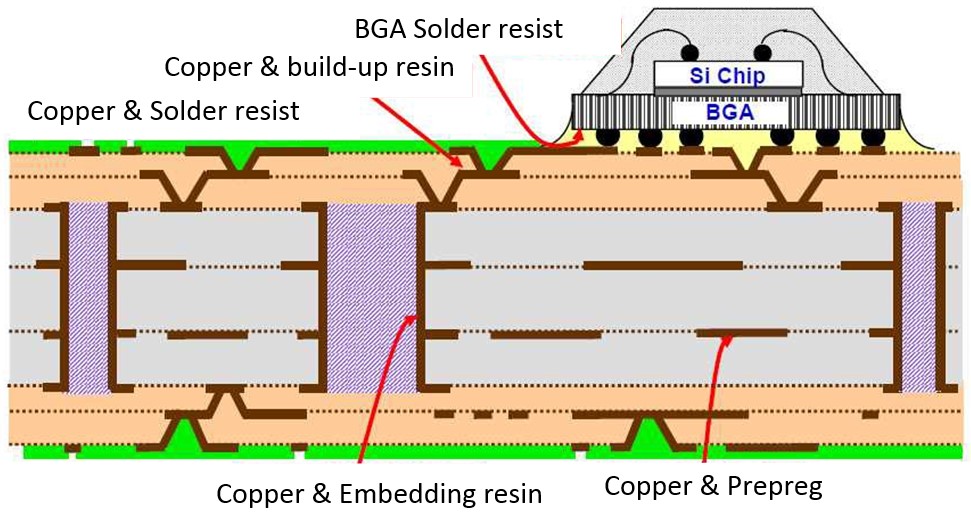}
\caption{
Illustration of a Printed Circuit Board. 
Copper circuits are made with various insulators for several purposes.  
}
\end{figure}

Since PCBs have been required to have higher heat-resistant properties in recent years, 
copper surface treatment technologies have performed a more and more important role in the manufacturing process. 
This is because they can enable PCBs to maintain high copper to resin adhesion even under harsh conditions. 
Copper surface treatment (copper surface roughening) offers one of the most effective ways to increase adhesion between copper and resin. 
Copper surface roughening has been widely used for the purpose of increasing adhesion of copper to resins. 
It produces a unique surface topography which enhances the mechanical bonding of copper to resins, as illustrated in Figure 2.

\begin{figure}[h]
\centering
\includegraphics[width=0.9\linewidth]{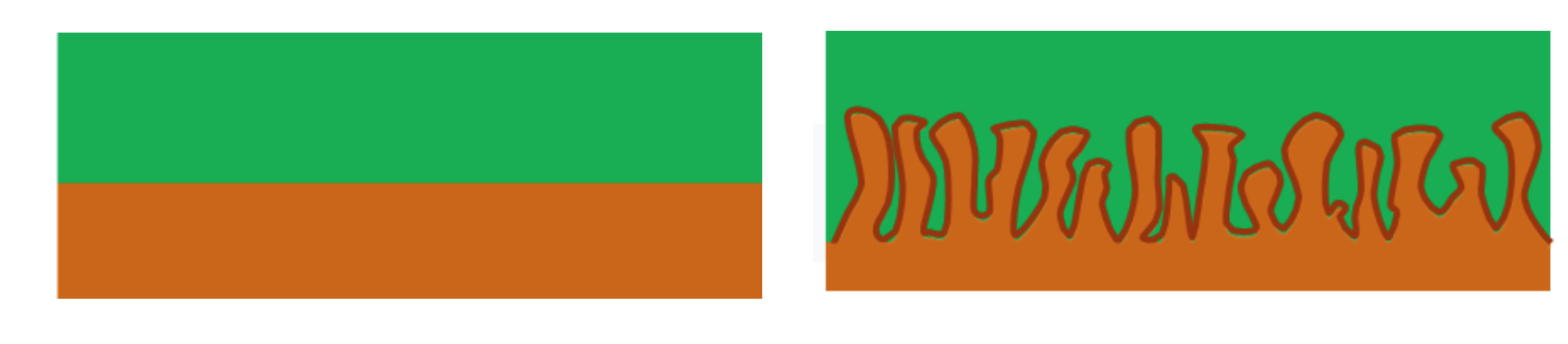}
\caption{
Illustration of a copper roughness surface. 
(Left) a perfectly smooth surface provides small adhesive surface 
as the interface between copper, in brown, and the resin, in green is minimal. 
(Right) a rough surface provides a larger surface at the interface of the resin. 
Larger contact surface areas provide higher adhesive potency.
}
\end{figure}

In very broad terms, rough surface allow for stronger bonds 
as the asperities of the surface provide a wide range of adhesive surface area and an anchoring effect. 
In contrast, smooth surfaces don’t provide such effects so that they have lower adhesive potential. 
In the remaining of this paper, we will refer to the potential bonding strength of a copper surface as its ``adhesive potency''.
It is also important to note that the adhesive potency of a copper surface is related to its ``roughness'', 
which is observable at the microscopic scale.

Electronic substrate manufacturers have developed advanced manufacturing processes to shape the surface of copper sheets in order to increase their adhesive potency. 
This is typically achieved by applying an etching solution on the copper surface.
Being able to accurately assess the adhesive potency of a copper surface would 
allow to further optimize manufacturing processes to increase the reliability of electronic devices.
However, assessing the adhesive potency of a copper surface is a complex task, even for the most expert practitioners. 
Hence the motivations of this study is two fold: 
First we aim to automate the evaluation of a copper sheet adhesive potency from microscopic imaging of its surface. 
Second, we aim to better understand the affect of different manufacturing processes on the adhesive potency of copper surfaces.
Towards this goal, we built a dataset of microscopic images of copper surfaces, which we detail below.

We imaged copper surfaces using Scanning Electron Microscopy (SEM) at a resolution of $100$ nm.
To investigate the impact of different manufacturing processes on the copper surface, 
we applied 16 different etching solutions, with decreasing etching power, to the copper surface.
For each of these solutions, we captured 50 SEM images of $960 \times 1280$ pixels so that the full dataset
consists of $800$ ($16 \times 50$) images.
Each image is annotated with a label $t$ corresponding to the etching solution used to shape the copper surface.
Each solution was obtained by submitting the original solution $t=0$ to an extreme stress test for a period of time $t$.
Hence, we know that for all images of copper surfaces with label $t$ show higher adhesive potency than the images labelled with $t' > t$. 
However, we do not know the \textit{exact} impact of the stress test on the surface adhesive potency.

\begin{figure}[h]
\centering
\includegraphics[width=0.9\linewidth]{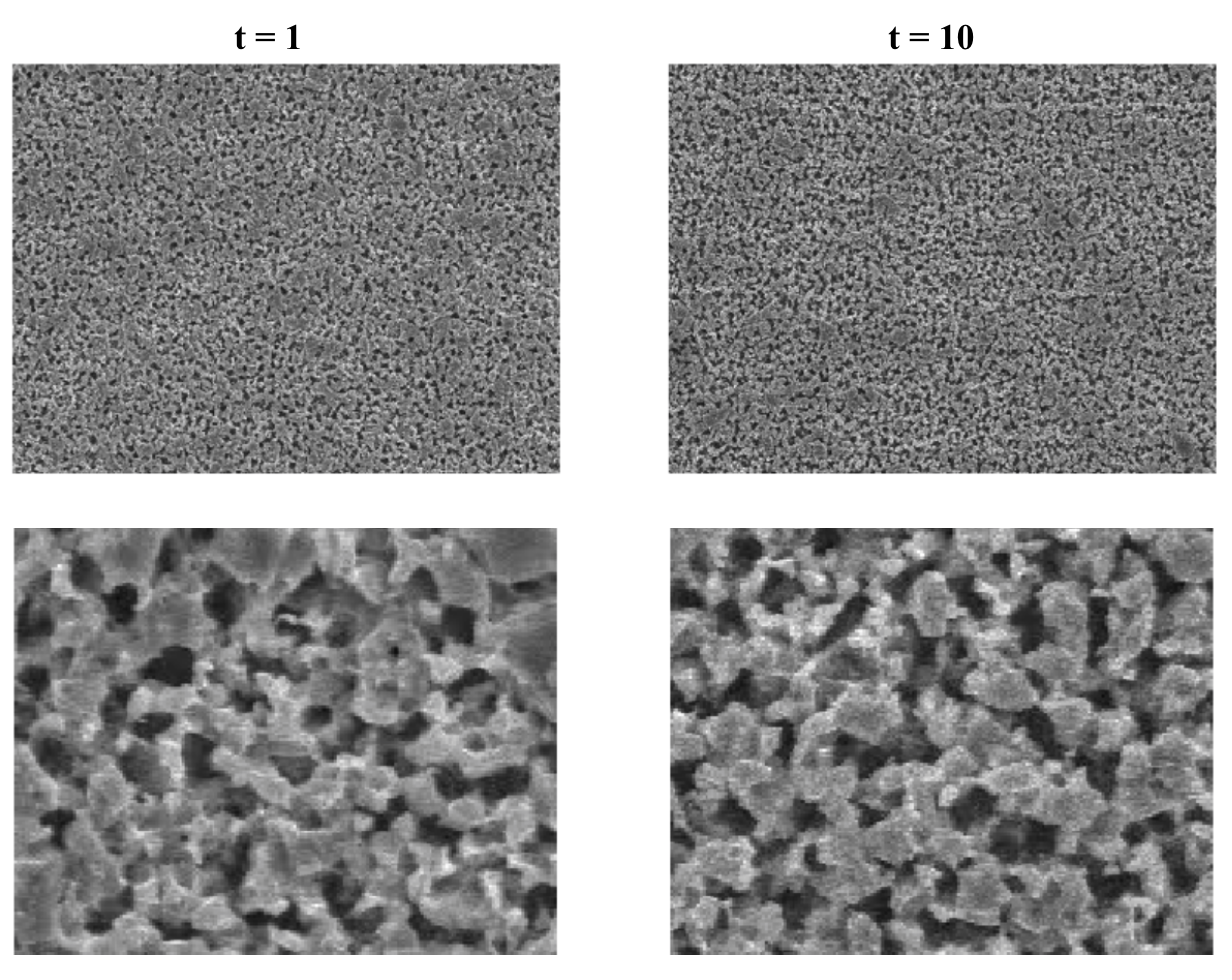}
\caption{
Illustration of a few images from the dataset. 
(Top) Full images. 
(Down) Zoomed-in areas of $200 \times 200$ pixels. 
(Right) Sample image of label $t=1$. 
(Left) Sample image of label $t=10$. 
The difference between both images are minimal to an untrained eye.
Precisely defining the visible difference with words is a difficult task.
}
\end{figure}

\section{Method}

\subsection{Dataset and Notations}
We denote the dataset described above as $\mathcal{D}=\{(x_i,t_i) | i \in [1,800] \}$ where $x$ denote gray scale images $x \in \mathbb{R}^{H \times W}$ and labels $t \in [0,16]$ corresspond to the time of stress test applied.
We split the dataset $\mathcal{D}$ into a training $\mathcal{D}_{tr}$, validation $\mathcal{D}_{val}$ and test $\mathcal{D}_{te}$ set so that the number of images $x$ per label $t$ in each set are 40 for training set, and 5 for the validation and test sets.

\subsection{Baseline}

We start by establishing a strong baseline for our study.
The baseline architecture follows standard convolutional network designs for image classification.
This architecture, illustrated in Figure 4, is made of several residual blocks 
sequentially interleaved with max pooling operations.
Each residual block consists of $n$ repetitions of a sequence of 
$3 \times 3$ Convolution, Batch Normalization and ReLU layers,
followed by a residual skip connection.
We set $N$ residual blocks between every max pooling layer
and we denote by $d$ the number of pooling layers.
Hence, the full depth $D$ of the network (in number of convolution layers) 
is given by $D=d \times n \times N +1$, where the term $1$ corresponds to the
initial $3 \times 3$ Convolution layer happening before the first pooling operation.
Finally, the top layer of the network is made of a global average pooling layer 
followed by a linear softmax layer with output dimension 15 corresponding to our number of classes.
For simplicity and contrary to standard practices, we keep the number of channels $c$ constant 
in all layers of the network.

\begin{figure}[h]
\centering
\includegraphics[width=0.9\linewidth]{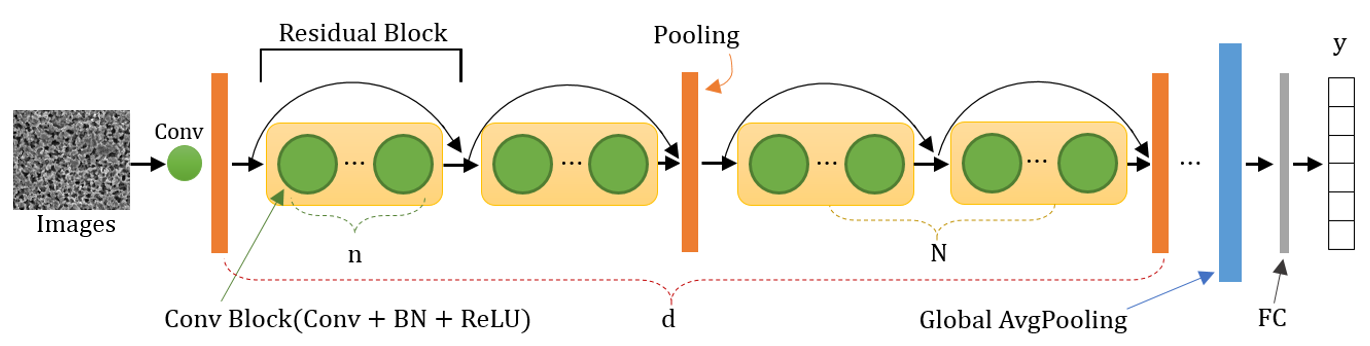}
\caption{
Illustration of our baseline architecture.
With our modular architecture definition, 
the architecture is fully defined by parameters $n$,$N$,$c$ and $d$.
}
\end{figure}

With this parameterization, our network is fully specified by the four hyper parameters $c$, $d$, $n$ and $N$.
We performed a grid search over these hyper parameters to select the best performing architecture.
The details of this architecture search are given in the experiment section,
and, as we shall see then, the best performing architecture performs significantly better than human experts.
However, the decision process through which this model reaches such a high accuracy is entirely opaque
as the distributed nature of the model's internal representations provides little interpretability.
The remainder of this section details our attempt to design an architecture that can provide 
useful explanations of the process through which high accuracy recognition can be performed.

\subsection{Assumptions}
The basic idea behind our method is similar to the visual insights provided by the
visualization method of \cite{yosinski2015understanding} and the attention mechanisms in visual models 
for image captioning \cite{you2016image} and question answering \cite{li2018knowing} tasks:
We would like to evaluate the contribution of each input pixel to the final classification decision.
Indeed, explicitly pointing out the surface regions responsible 
for low and high adhesive potency would provide guidance for human experts to 
identify visual patterns characteristic of either cases.

To do so, we modify our initial problem formulation into a segmentation task:
Given an input $x \in \mathbb{R}^{H \times W}$, we want to design a model that outputs 
a segmentation mask $s \in \mathbb{R}^{H \times W}$ assessing the contribution of each individual pixel to the output adhesive potency score (i.e., the output class $t$).
Training a typical segmentation model for this task would require ground truth 
segmentation masks $s$ for each image $x$ of the training set.
However, manually annotating ground truth segmentation masks for this task is not feasible, 
as human experts are not able to provide such fine-grained annotations.
Instead, we have to train the segmentation given a single ground truth label $t$ per image.
In order to do so, we make the following assumption:

\textbf{Assumption}: 
For a given image $x$ with label $t$, 
the \textit{pixel-wise} values of the true (unknown) segmentation mask $s$
follow a spatially stationary binomial distribution whose expected value, 
averaged over the spatial dimensions of $x$, is given by $t$ following:

\begin{equation}
  s_{hw} \sim \mathcal{B}(f(t)), \forall h,w,t \in H \times W \times T
\end{equation}

in which we introduced a target function $f$, which we shall discuss in Section 3.6.

In other words, we suppose that there exists a true binary segmentation map
$s$ assigning to each individual pixel of $x$ a binary adhesive potency score:
$s$ takes 0 values in regions of the surface providing low adhesive potency (i.e. smooth copper surface areas)
and 1 values in regions of the surface providing high adhesive potency (i.e. rough copper surface areas).
The adhesive potency score of an entire image $x$ is thus given by the average of the pixel-wise values,
and this average value is uniquely defined by the image label $t$.

\begin{figure}[h]
\centering
\includegraphics[width=0.9\linewidth]{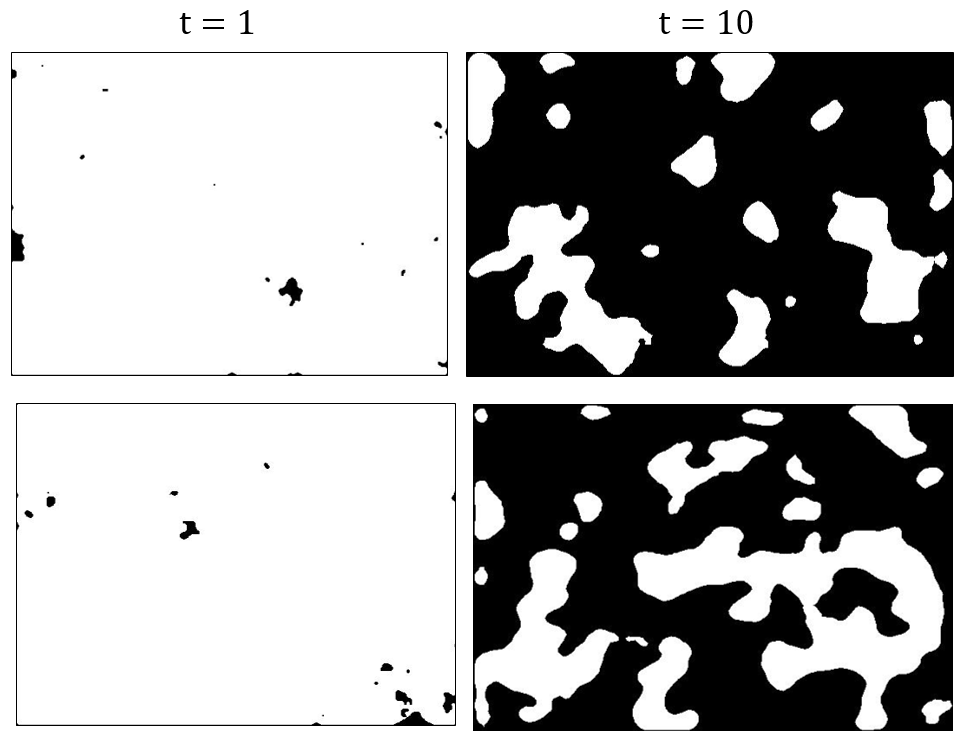}
\caption{
Illustration of binary segmentation masks. 
White pixels represent areas of high adhesive potency
and black surface represent areas of low adhesive potency.
We make the assumption that the ratio of white surface is 
constant for different samples (top and bottom) with equal label $t$.
The exact value of this ratio is defined by the target function $f$.
}
\end{figure}

Figure 5 provides a visual illustration of this idea.
For each image, we suppose the existence of such binary mask $s$
quantifying the adhesive potency of local areas of the surface.
The average value of the binary mask $s$ amounts to the ratio of the copper surface covered in white 
(i.e. with value 1, corresponding to high adhesive potency).
Our assumption means that this ratio stays constant for different 
images $x$ sharing a similar label $t$,
and the exact value of this ratio is given by the target function $f$.

\subsection{Architecture}

%
In this section, we present the architecture used 
to compute binary segmentation masks from input images.
Our architecture extends the baseline architecture presented in Section 3.1,
with an ascending path that progressively upsamples the output
of the descending path, similar to the UNet architecture \cite{ronneberger2015u} and illustrated in Figure 6.

Residual modules of the ascending path are the exact symmetric of their analog in descending path.
The ascending path uses bilinear upsampling layers 
instead of the max pooling layers of the descending path.
Different from the UNet architecture, and following previous works \cite{lee2017superhuman,smith2016deep},
we merge the outputs of the descending path modules with the inputs of the ascending path 
by summation, instead of concatenation. 
We also use valid convolutions to preserve the spatial resolution of the output.
Finally, we add a sigmoid layer at the top of the network in order to bound the output values between 0 and 1.

\begin{figure}[h]
\centering
\includegraphics[width=0.9\linewidth]{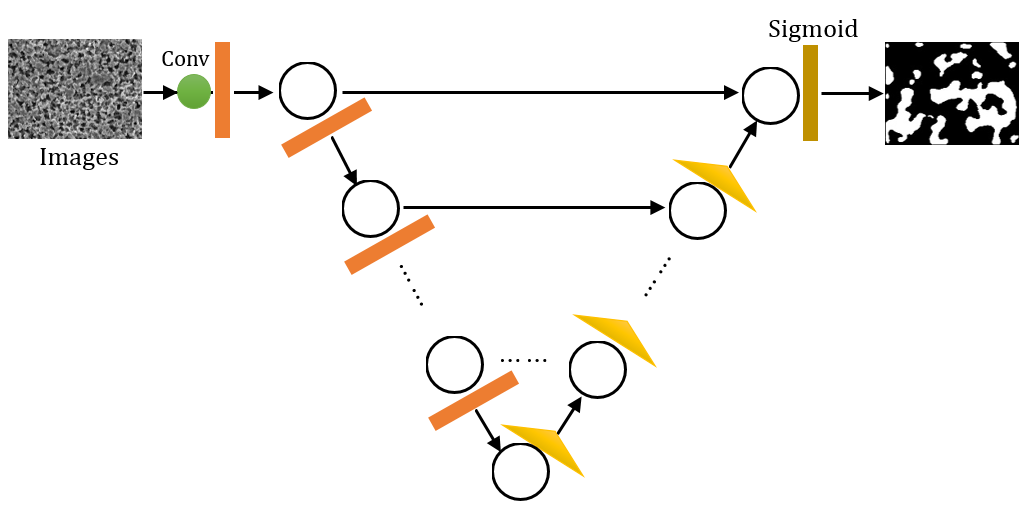}
\caption{
Illustration of the segmentation model architecture.
This architecture follows standard practice in 
UNet-like segmentation architectures.
}
\end{figure}

\subsection{Loss Function}

Given a segmentation model $M_{\theta}$, 
with weight parameters $\theta$, 
and an input image $x$,
we denote the average of the model output by $m_{\theta}(x)$:

\begin{subequations}
\begin{align}
M_{\theta}&: \mathbb{R}^{H \times W} \rightarrow [0,1]^{H \times W} \\
\tilde{s}&= M_{\theta}(x) \\
m_{\theta}&:  \mathbb{R}^{H \times W} \rightarrow [0,1]\\
m_{\theta}(x)&=  \frac{1}{HW} \sum_{h,w} \tilde{s}_{hw}
\end{align}
\end{subequations}

We can then train the model by regressing the average value of the segmentation mask to the
target label given by $f(t)$. 
Given a training dataset of labeled samples $\mathcal{D}_{tr}={(x_i, t_i)}$,
learning is thus done by minimizing the following loss function over the model's parameters $\theta$:

\begin{equation}
\theta^* = argmin_{\theta} \mathbb{E}_{x,t \in \mathcal{D}_{tr}} ||m_{\theta}(x) - f(t) ||^2 
\end{equation}

However, as we shall see in the next section, 
the target function $f$ represents an unknown ideal function,
so we do not have access to the actual values of $f(t)$.
Instead, we will approximate $f$ with a known hypothesis function $g \approx f$,
so that the actual training loss used in our experiments is:

\begin{equation}
\theta^* = argmin_{\theta} \mathbb{E}_{x,t \in \mathcal{D}_{tr}} ||m_{\theta}(x) - g(t) ||^2 
\end{equation}

\subsection{Target Function}

In section 3.3, we have made the assumption 
that labels $t$ uniquely define the expected value 
of the ground-truth binary segmentation masks $s$ through a target function $f$.
In this section, we discuss the role of this target function.

The target function $f(t)$ describes the evolution of the copper surface adhesive potency with time $t$.
More precisely $f$ defines the evolution of the \textit{ratio of adhesive surface area} with time (see Figure 5).

However, $f$ is an ideal, unknown function, of which we have only supposed the existence.
We do not know the value taken by $f(t)$ for a given $t$ because we 
do not know the exact impact of the manufacturing process on the copper's surface characteristics.
We thus introduce a known hypothesis function $g$ to approximate the ideal target function $f$.
$g$ expresses our belief of what the values taken by the true function $f(t)$ are.
Although we do not know the exact values taken by $f$, we know several of its characteristics,
which we can use to reduce the search space of hypothesis functions $g$:

In Section 2, we have established that, for all $t$, 
copper surfaces with label $t' > t$ should have lower adhesive potency than copper surfaces of label $t$.
Hence, $f$ should be a monotonically decreasing function of time,
and we can thus restrict our search of hypothesis function $g$ to monotonically decreasing function.
For example, the simplest of such function would be the linear function, 
taking linearly decreasing values from 1 to 0,
which we start by analyzing in the experiment section.

Second, as $f$ is assumed to describe the true evolution of the copper surface statistics with $t$,
a universal function approximator should be able to perfectly learn its values, given enough training data.
Hence training an ideal 

training an ideal model with the supervision signal provided by $f$ should yield low test errors.
Hence training a segmentation model with a hypothesis $g \approx f$ should lead to low test errors.
This means that we can use the error of the model on the held out validation dataset as a proxy measure on
the validity of the hypothesis function $g$.

In the experiment section, we evaluate the model errors with different hand-crafted hypothesis functions $g$,
and show that the model trained on a poorly chosen hypothesis function $g$ (i.e. non monotonically decreasing hypothesis) 
tend to yield large test errors, confirming our idea that generalization error might be used as proxy metric
to quantify the validity of a hypothesis function $g$.
In future work, we aim to jointly learn the hypothesis function $g$ with the model parameters $\theta$ to automatically
formulate hypothesis

\section{Related Work}

We identify three different lines of research that share similarities with our study:
Explainability of learned visual representations, 
weak supervision of segmentation models and 
machine learning applications for material science.
Our approach can be seen as a weakly supervised approach to provide material 
experts with strong supports for interpretable decision, 
which falls at the intersection of these three research lines.
We briefly present some of these works in the following subsections

\subsection{CNN Interpretability}
The actual processing performed by deep models is hard to interpret:
because of the distributed nature of the model's internal representation,
it is difficult to assign a useful meaning to each individual unit meaning.
This is problematic as model failure cases are very hard to investigate and justify.
Hence, an interesting line of work is researching for tools to interpret neural network processes.
We can see two approaches in this line of work:
one is focused on visualization and the other one focused on generating natural language explanations.
 
On the side of natural language explainability, Park \textit{et al.} \cite{park2016attentive} jointly trained to answer and justify for their
answers on a visual question answering task. 
They combine visual attention maps and natural language generation to bring interpretability to the model's output.
 
More related to our work is the line of work focused on visualizing neural network hidden activations
Yosinski \textit{et al.} \cite{yosinski2015understanding} provide two useful tools for visualizing and interpreting neural nets. 
One is to visualize  the activations produced on each layer of a trained convnet as it 
processes an image or video, and the other enables visualizing 
features at each layer of a DNN via regularized optimization in image space.

More recently, Carter \textit{et al.} \cite{carter2019activation} propose an explorable activation atlas of vision model's learned feature by using feature inversion to visualize millions of activations from an image classification network
Their technique provides insights regarding the network's conceptual representations.

Visual attention \cite{xu2015show} also helps in explaining the process through which model's 
outputs are computed by providing visual clues as to what regions of the input space 
contribute the most to the final decision. 
These have been used to investigate model's operations on image captioning \cite{xu2015show,you2016image} and VQA \cite{antol2015vqa,zhang2016yin,goyal2017making} tasks,
and is similar to the idea behind our segmentation model.

\subsection{Weakly Supervised Segmentation}
An other line of our research is based on the idea of pixel-wised segmentation with weak supervision.
Most deep learning methods for computer vision tasks such as object recognition\cite{simonyan2014very,russakovsky2015imagenet} and object detection (\textit{e.g.} DetectorNet \cite{szegedy2013deep}, OverFeat \cite{sermanet2013overfeat}, R-CNN\cite{girshick2014rich}, SPP-net\cite{he2015spatial}, Fast R-CNN\cite{girshick2015fast}, and Faster R-CNN\cite{ren2015faster}) rely on strongly annotated data and these successful techniques require ground-truth labels to be given for a big training data set, 
however it is an expensive and time-consuming effort to attain strong supervison information due to the high cost of data annotation.

Weakly supervised methods\cite{bilen2016weakly,peyre2017weakly,arandjelovic2016netvlad} have shown great promise while noisy, limited, or imprecise sources are used to provide supervision signal for labeling large amounts of training data in a supervised learning setting, that means the level of annotation can be less detailed. 
There are several weakly-supervised studies\cite{bilen2014weakly,bilen2015weakly,song2014learning} in CNN-based object detection, adopting CNN as a feature extrator and using training images with only image-level labels and no bounding boxes.

Moreover, Eugene Vorontsov\textit{et al.} \cite{vorontsov2019boosting} develop a segmentaiton method with weakly supervision by image-to-image translation between weak labels.
Yuxing Tang \textit{et al.}\cite{tang2016large} build a similarity-based knowledge transfer mode trying to investigate whether knowledge about visual and semantic similarities of object categories can help improve the peformance of detectors trained in a weakly supervised setting.

The most related work with ours is Rihuan Ke's\cite{ke2019multi} in which they present a weakly-supervised learning strategy for segmentation with lazy labels and develop a multi-task learning framework to integrate the instance detection, separation and segmentation within a deep neural network.

\subsection{Machine Learning for Material Science}
Deep learning has showed remarkable success on many important 
learning problems in chemistry\cite{goh2017deep}, 
drug discovery\cite{gawehn2016deep,chen2018rise}, 
biology\cite{webb2018deep} and materials science.
In the field of materials science, deep neural networks have also been receiving increasing attention and have achieved great improvements, for example, in material property prediction and new materials discovery for batteries\cite{pilania2013accelerating,hansen2015machine}\cite{ward2016general,liu2017materials}.
CNNs have also been used for detect analysis on microscopic images of various material surfaces\cite{lubbers2017inferring,li2018automated,nash2018review}.

Beyond the material sciences, we note a growing interest in applying deep learning techniques
for scientific discovery. 
This perhaps best exemplified by the impressive successes of AlphaFold \cite{evans2018novo} in protein folding estimation,
or the Celeste \cite{regier2015celeste} project which cataloged celestial objects of visible universe.

Our work, while much more modest in its scale, 
shares the characteristic of using vision models to unravel
the underlying principle of material reactions to chemical treatments.

\section{Experiments}

\subsection{Classification results}
We start by evaluating the baseline classification model described in Section 3.3 and illustrated in Figure 4.
The hyper parameters $n$, $N$, $d$, and $c$ of the classifier architecture 
were obtained by a grid search within the limits of a 12GB Nvidia GPU memory size .
The model was trained on the training dataset for 200 epochs using the Adam optimizer with default parameterization.

We compare the classification outputs of the baseline model to those 
of an expert material scientist on a blind test.
The human expert was given the sample images of the training set to practice,
and we evaluated the expert's answers on the samples of the test set.

\begin{figure}[h]
\centering
\includegraphics[width=0.9\linewidth]{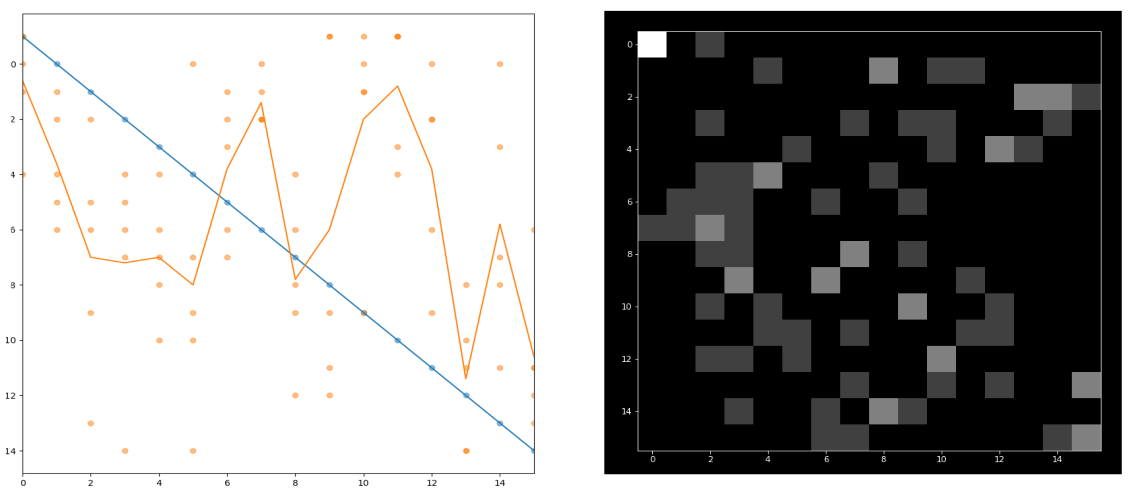}
\caption{
Results of the human evaluation. 
(Left) Regression plot on the test set. 
The blue line illustrates the ground-truth labels.
Each yellow dot represents the expert's predicted labels on individual images. 
The yellow line represents the expert's average answers for images of similar ground-truth label.
(Right) Expert's answers visualized as a confusion matrix.
}
\end{figure}

Figure 7 shows the results achieved by the model and Figure 8 shows the results achieved by our baseline model.
As can be seen in these figures, the model tends to predict the manufacturing 
process more accurately than the human evaluator.
However, these results should be taken with a grain of caution 
as the human expert was given little time to practice on this specific dataset,
while the model was selected as the best performing baseline from an extensive parameter search.
We plan on re-conducting the human evaluation in an updated version of this paper.

\begin{figure}[h]
\centering
\includegraphics[width=0.9\linewidth]{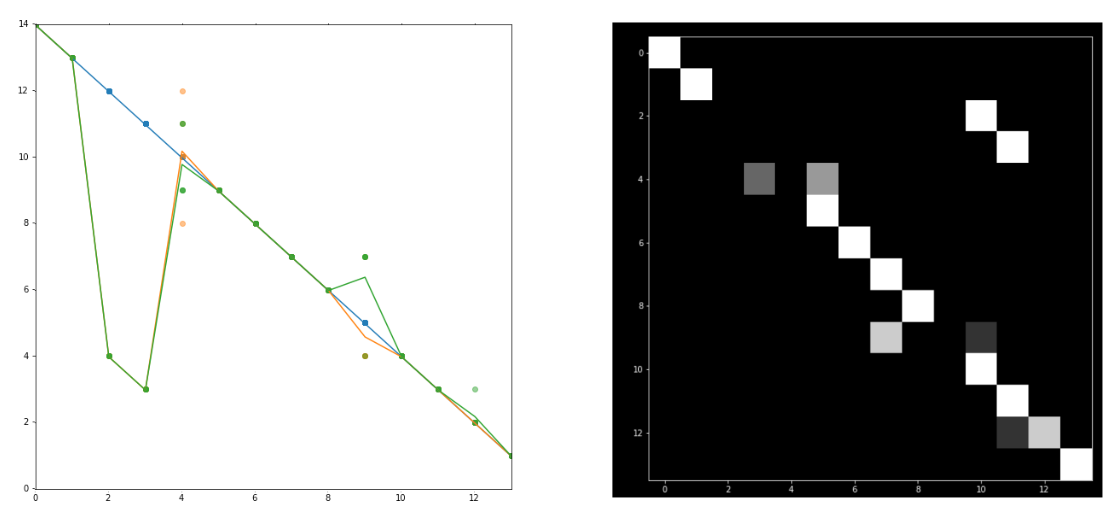}
\caption{
Results of the model evaluation.
(Left) Regression plot. 
Blue dots represent predictions on the training samples, 
yellow dots represent validation samples and green dots represent the test samples.
(Right) Confusion matrix of the test samples
}
\end{figure}

Interestingly, however, the human expert seems to accurately recognize 
the surfaces of highest adhesive potency as seen in the downward trend of his results 
for low $t$ values. For example, he easily identified the surface for $t=0$.
On the other hand, his guesses for low adhesive potency are much more random.

This is in stark contrast with the classifier accuracy, which perfectly identifies
the low adhesive potency surfaces (high $t$ values), but consistently misclassifies 
surfaces of high adhesive potency (i.e.; for $t=2$ and $t=3$).

\subsection{Visualization of Segmentation Results}

To motivate our results, Figure 9 illustrates an output of the segmentation model on a small patch of an input image $x$.
To a novice observer, the model seems to assign lower adhesive potential to smoother regions of the input.
In future work, we plan on further exploring these visualizations with human experts to better understand the 
patterns characterizing the surface adhesive potency.

\begin{figure}[h]
\centering
\includegraphics[width=0.9\linewidth]{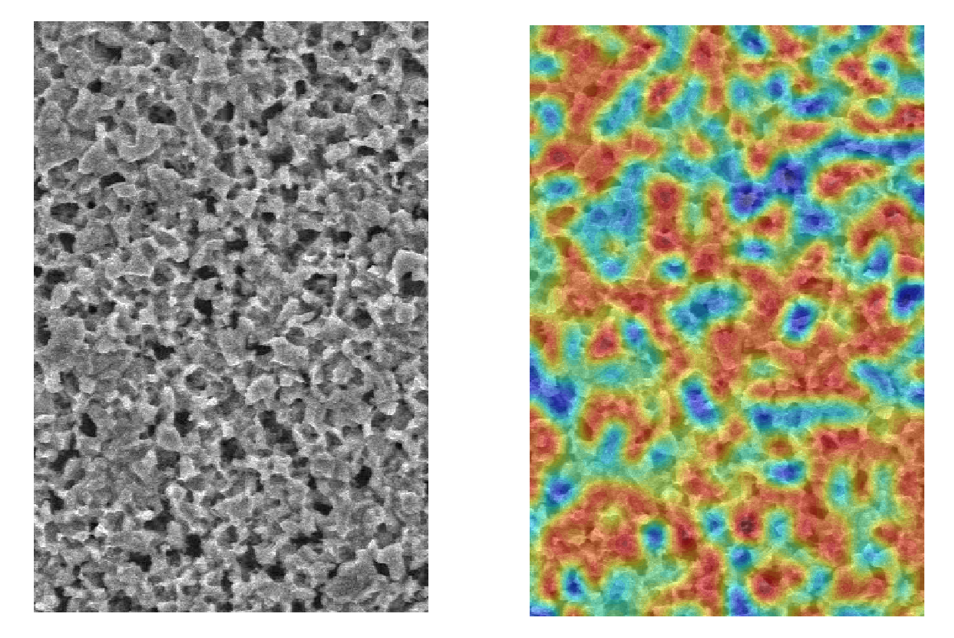}
\caption{
Visualization of the segmentation model output.
(Left): Zoomed-in input image.
(Right): Model segmentation output overlaid on the input image.
Blue areas represent 1 values (high adhesive potency) 
and red areas represent 0 values (low adhesive potency).
}
\end{figure}

\subsection{Investigation of the hypothesis function}
In section 3.6, we made the very informal argument that the model's test error could be used
to evaluate the validity of different hypothesis $g$ describing the evolution of
the copper surface characteristics with time $t$. 
In this section, we attempt to experimentally validate this idea.

We start by evaluating the linear function $g(t)=1 - \frac{t}{T}$.
This hypothesis function formulates the idea that for $t=1$, 
all the  segmentation mask values should be $1$, 
so that the whole surface area should have strong adhesive potency, 
while for $t=14$, all the segmentation mask values should be $0$, so that 
the whole surface area should have low adhesive potency.
For each $t$ between 1 and 14, this hypothesis function defines a linear decrease
of the low adhesive potency surface ratio.

\begin{figure}[h]
\centering
\includegraphics[width=0.9\linewidth]{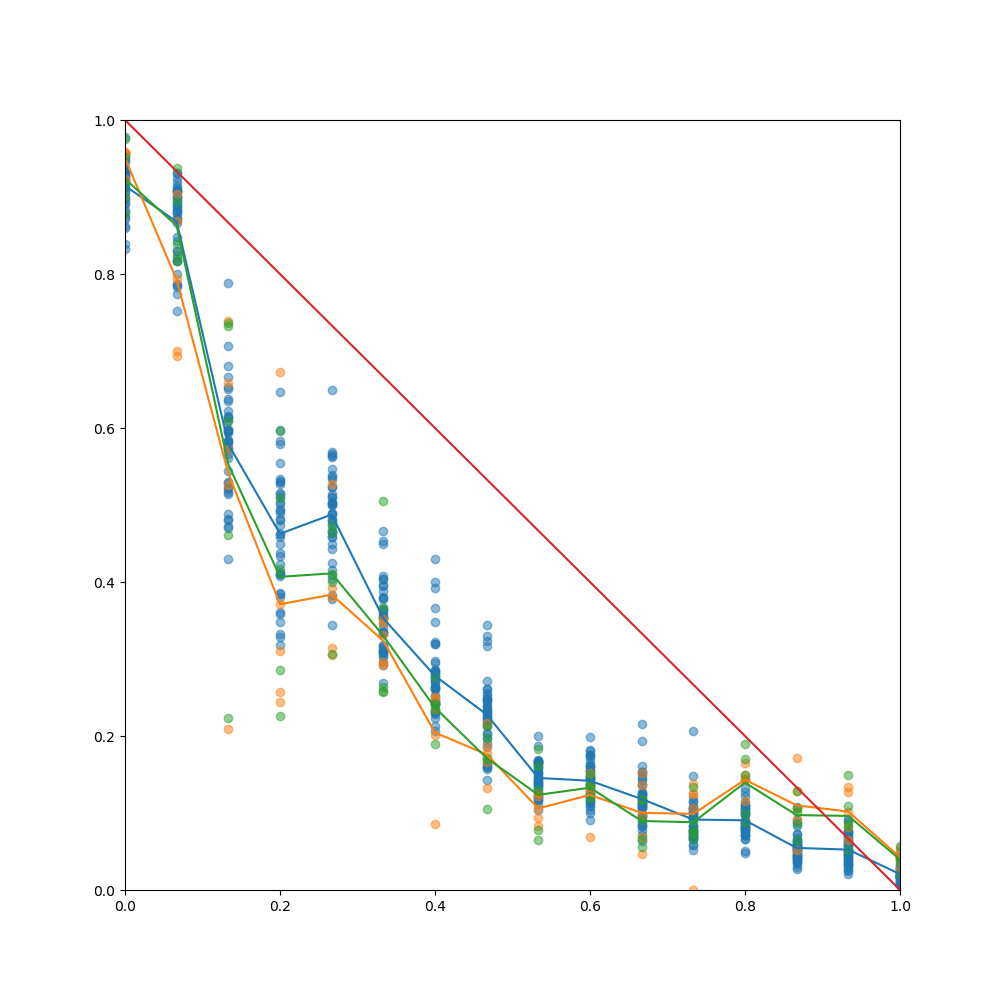}
\caption{
Results of the segmentation model evaluation.
The red line illustrates the linear hypothesis used for training.
Blue, yellow and green dots represent individual image output predictions 
on the training, validation and test set respectively.
Even on training samples, the model is not able to overfit the ground labels 
generated by the hypothesis function.
Instead of a linear decay, the model seems to predict a more aggressive decay,
following a square root or logarithmic trend.  
}
\end{figure}

\begin{figure*}[h!]
\centering
\includegraphics[width=0.9\linewidth]{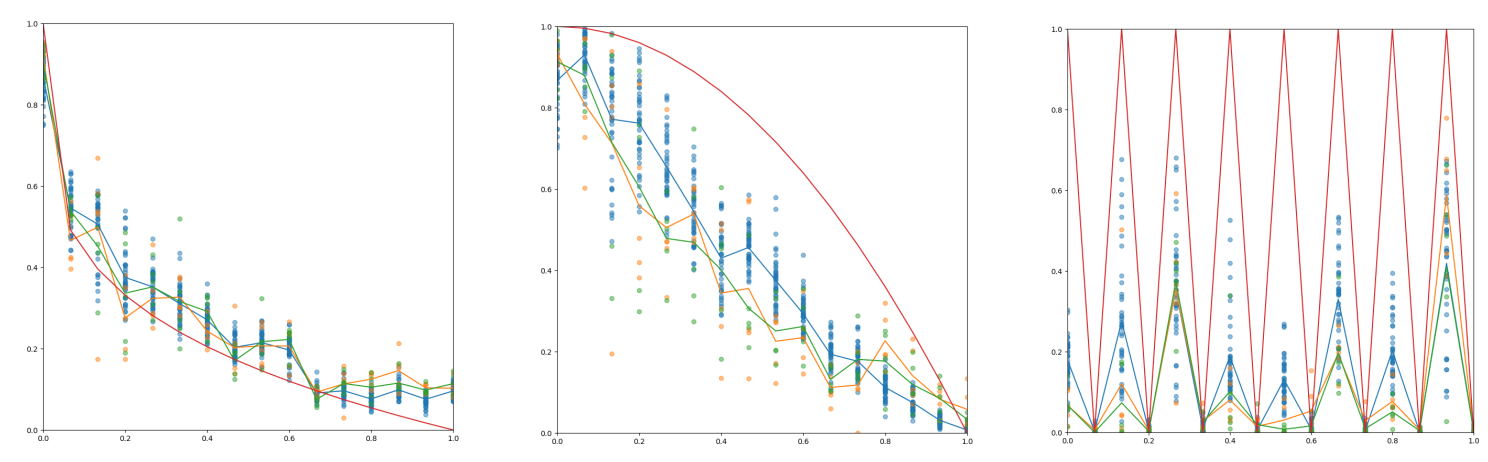}
\caption{
Results of the segmentation model evaluation with different hypothesis function $g$.
(Left) $g=1-t^{-\frac{1}{2}}$. The model seems to nicely fit the hypothesis function.
(Middle) $g=1-t^{-2}$. The model diverges from the hypothesis function, suggesting a more rapid decay.
(Right) $g=sign(-1^{t})$. The model struggles to learn the hypothesis function, yielding high errors.
}
\end{figure*}

Figure 10 compares the hypothesis function to the actual model output on the training, 
validation and test set.
First, we observe no overfitting as all datasets yield similar average results.
Second, we observe that instead of following a linear trend from 0 to 1, 
the surface ratio seems to evolve more similarly to a squared root or logarithmic function
with a sharp decrease in value for low $t$ and a slower decrease in higher values of $t$.
It is interesting to note that the same behavior is also observed on samples of the training set, 
for which the model was explicitly trained to reproduce the linear hypothesis function.
This result suggests that the surface ratio 
(i.e. the smoothing effect of the stress test on the copper surface) 
may decrease sub-linearly with time.

Motivated by this first experiment, 
we train the model on different hypothesis functions $g$,
illustrated in Figure 11.
We find that training a model with a hypothesis functions $g$ as suggested by our first experiment
yield low model errors, while training the model with an unrealistic hypothesis (i.e. a sine function)
yield high errors. 

\section{Discussion \& Conclusion}

In this paper, we have argued that deep neural networks can be used to 
help explain the process behind low-level recognition tasks.
We have focused on the task of assessing the adhesive potency 
of copper surfaces in the context of PCB manufacturing.

This is an interesting task to showcase the limitations of human's ability to explain their visual process
as it is a very low level recognition task for which trained experts can provide qualitative guesses while
not being able to fully justify for their guess.

On this task, we have first shown that an unrestricted classifier 
architecture can outperform a human expert in accuracy.
However, this architecture does not bring us any insight 
into \textit{why} a given copper surface should be classified 
as a high or low adhesive potency.

To shed some light into the decision process of the model,
we proposed to cast this problem as a a segmentation problem, 
in which the model is tasked with assigning a binary score to each pixel,
indicating whether the local area represented by this pixel provides a high or low adhesive power.
Visualizing the segmentation output of the model may prove useful for 
human experts to better understand the characteristics of high and low 
adhesive potency surfaces.

Finally, we developed a weak label training procedure to train this segmentation model.
Our procedure relies on a hypothetical relationship between the manufacturing process and the
copper surface roughness.

We very informally argued that the model error may reflect the validity of the hypothesis function,
and provided preliminary experiment results that tend to confirm our informal argument:
we observe lower errors when training the model with hypothesis functions that \textit{seem} more plausible,
and lower model error on unrealistic hypothesis functions (i.e. a sine function).

However, the result of this last experiment should be taken with caution as this study is still in a very preliminary stage.
In particular, our training procedure relies several assumptions that most likely do not hold in reality: 
In particular, we assumed that each individual area could be represented by a single binary value defining its adhesive potency
while, in reality, the adhesive potency of local areas are most likely not binary in nature.

Nevertheless, these preliminary results are encouraging, and we will continue our analysis in future work.

{\small
	\bibliographystyle{ieee}
	\bibliography{egbib}

\begin{thebibliography}{10}\itemsep=-1pt

\bibitem{antol2015vqa}
S.~Antol, A.~Agrawal, J.~Lu, M.~Mitchell, D.~Batra, C.~Lawrence~Zitnick, and
  D.~Parikh.
\newblock Vqa: Visual question answering.
\newblock In {\em Proceedings of the IEEE international conference on computer
  vision}, pages 2425--2433, 2015.

\bibitem{arandjelovic2016netvlad}
R.~Arandjelovic, P.~Gronat, A.~Torii, T.~Pajdla, and J.~Sivic.
\newblock Netvlad: Cnn architecture for weakly supervised place recognition.
\newblock In {\em Proceedings of the IEEE conference on computer vision and
  pattern recognition}, pages 5297--5307, 2016.

\bibitem{bilen2014weakly}
H.~Bilen, M.~Pedersoli, and T.~Tuytelaars.
\newblock Weakly supervised object detection with posterior regularization.
\newblock In {\em British Machine Vision Conference}, volume~3, 2014.

\bibitem{bilen2015weakly}
H.~Bilen, M.~Pedersoli, and T.~Tuytelaars.
\newblock Weakly supervised object detection with convex clustering.
\newblock In {\em Proceedings of the IEEE Conference on Computer Vision and
  Pattern Recognition}, pages 1081--1089, 2015.

\bibitem{bilen2016weakly}
H.~Bilen and A.~Vedaldi.
\newblock Weakly supervised deep detection networks.
\newblock In {\em Proceedings of the IEEE Conference on Computer Vision and
  Pattern Recognition}, pages 2846--2854, 2016.

\bibitem{carter2019activation}
S.~Carter, Z.~Armstrong, L.~Schubert, I.~Johnson, and C.~Olah.
\newblock Activation atlas.
\newblock {\em Distill}, 4(3):e15, 2019.

\bibitem{chen2018rise}
H.~Chen, O.~Engkvist, Y.~Wang, M.~Olivecrona, and T.~Blaschke.
\newblock The rise of deep learning in drug discovery.
\newblock {\em Drug discovery today}, 23(6):1241--1250, 2018.

\bibitem{evans2018novo}
R.~Evans, J.~Jumper, J.~Kirkpatrick, L.~Sifre, T.~Green, C.~Qin, A.~Zidek,
  A.~Nelson, A.~Bridgland, H.~Penedones, et~al.
\newblock De novo structure prediction with deeplearning based scoring.
\newblock {\em Annu Rev Biochem}, 77:363--382, 2018.

\bibitem{gawehn2016deep}
E.~Gawehn, J.~A. Hiss, and G.~Schneider.
\newblock Deep learning in drug discovery.
\newblock {\em Molecular informatics}, 35(1):3--14, 2016.

\bibitem{girshick2015fast}
R.~Girshick.
\newblock Fast r-cnn.
\newblock In {\em Proceedings of the IEEE international conference on computer
  vision}, pages 1440--1448, 2015.

\bibitem{girshick2014rich}
R.~Girshick, J.~Donahue, T.~Darrell, and J.~Malik.
\newblock Rich feature hierarchies for accurate object detection and semantic
  segmentation.
\newblock In {\em Proceedings of the IEEE conference on computer vision and
  pattern recognition}, pages 580--587, 2014.

\bibitem{goh2017deep}
G.~B. Goh, N.~O. Hodas, and A.~Vishnu.
\newblock Deep learning for computational chemistry.
\newblock {\em Journal of computational chemistry}, 38(16):1291--1307, 2017.

\bibitem{goyal2017making}
Y.~Goyal, T.~Khot, D.~Summers-Stay, D.~Batra, and D.~Parikh.
\newblock Making the v in vqa matter: Elevating the role of image understanding
  in visual question answering.
\newblock In {\em Proceedings of the IEEE Conference on Computer Vision and
  Pattern Recognition}, pages 6904--6913, 2017.

\bibitem{hansen2015machine}
K.~Hansen, F.~Biegler, R.~Ramakrishnan, W.~Pronobis, O.~A. Von~Lilienfeld,
  K.-R. Müller, and A.~Tkatchenko.
\newblock Machine learning predictions of molecular properties: Accurate
  many-body potentials and nonlocality in chemical space.
\newblock {\em The journal of physical chemistry letters}, 6(12):2326--2331,
  2015.

\bibitem{he2015spatial}
K.~He, X.~Zhang, S.~Ren, and J.~Sun.
\newblock Spatial pyramid pooling in deep convolutional networks for visual
  recognition.
\newblock {\em IEEE transactions on pattern analysis and machine intelligence},
  37(9):1904--1916, 2015.

\bibitem{ke2019multi}
R.~Ke, A.~Bugeau, N.~Papadakis, P.~Schuetz, and C.-B. Sch{\"o}nlieb.
\newblock A multi-task u-net for segmentation with lazy labels.
\newblock {\em arXiv preprint arXiv:1906.12177}, 2019.

\bibitem{lee2017superhuman}
K.~Lee, J.~Zung, P.~Li, V.~Jain, and H.~S. Seung.
\newblock Superhuman accuracy on the snemi3d connectomics challenge.
\newblock {\em arXiv preprint arXiv:1706.00120}, 2017.

\bibitem{li2018automated}
W.~Li, K.~G. Field, and D.~Morgan.
\newblock Automated defect analysis in electron microscopic images.
\newblock {\em npj Computational Materials}, 4(1):36, 2018.

\bibitem{li2018knowing}
W.~Li, Z.~Yuan, X.~Fang, and C.~Wang.
\newblock Knowing where to look? analysis on attention of visual question
  answering system.
\newblock In {\em Proceedings of the European Conference on Computer Vision
  (ECCV)}, pages 0--0, 2018.

\bibitem{liu2017materials}
Y.~Liu, T.~Zhao, W.~Ju, and S.~Shi.
\newblock Materials discovery and design using machine learning.
\newblock {\em Journal of Materiomics}, 3(3):159--177, 2017.

\bibitem{lubbers2017inferring}
N.~Lubbers, T.~Lookman, and K.~Barros.
\newblock Inferring low-dimensional microstructure representations using
  convolutional neural networks.
\newblock {\em Physical Review E}, 96(5):052111, 2017.

\bibitem{nash2018review}
W.~Nash, T.~Drummond, and N.~Birbilis.
\newblock A review of deep learning in the study of materials degradation.
\newblock {\em npj Materials Degradation}, 2(1):37, 2018.

\bibitem{park2016attentive}
D.~H. Park, L.~A. Hendricks, Z.~Akata, B.~Schiele, T.~Darrell, and M.~Rohrbach.
\newblock Attentive explanations: Justifying decisions and pointing to the
  evidence.
\newblock {\em arXiv preprint arXiv:1612.04757}, 2016.

\bibitem{peyre2017weakly}
J.~Peyre, J.~Sivic, I.~Laptev, and C.~Schmid.
\newblock Weakly-supervised learning of visual relations.
\newblock In {\em Proceedings of the IEEE International Conference on Computer
  Vision}, pages 5179--5188, 2017.

\bibitem{pilania2013accelerating}
G.~Pilania, C.~Wang, X.~Jiang, S.~Rajasekaran, and R.~Ramprasad.
\newblock Accelerating materials property predictions using machine learning.
\newblock {\em Scientific reports}, 3:2810, 2013.

\bibitem{regier2015celeste}
J.~Regier, A.~Miller, J.~McAuliffe, R.~Adams, M.~Hoffman, D.~Lang, D.~Schlegel,
  and M.~Prabhat.
\newblock Celeste: Variational inference for a generative model of astronomical
  images.
\newblock In {\em International Conference on Machine Learning}, pages
  2095--2103, 2015.

\bibitem{ren2015faster}
S.~Ren, K.~He, R.~Girshick, and J.~Sun.
\newblock Faster r-cnn: Towards real-time object detection with region proposal
  networks.
\newblock In {\em Advances in neural information processing systems}, pages
  91--99, 2015.

\bibitem{ronneberger2015u}
O.~Ronneberger, P.~Fischer, and T.~Brox.
\newblock U-net: Convolutional networks for biomedical image segmentation.
\newblock In {\em International Conference on Medical image computing and
  computer-assisted intervention}, pages 234--241. Springer, 2015.

\bibitem{russakovsky2015imagenet}
O.~Russakovsky, J.~Deng, H.~Su, J.~Krause, S.~Satheesh, S.~Ma, Z.~Huang,
  A.~Karpathy, A.~Khosla, M.~Bernstein, et~al.
\newblock Imagenet large scale visual recognition challenge.
\newblock {\em International journal of computer vision}, 115(3):211--252,
  2015.

\bibitem{sermanet2013overfeat}
P.~Sermanet, D.~Eigen, X.~Zhang, M.~Mathieu, R.~Fergus, and Y.~LeCun.
\newblock Overfeat: Integrated recognition, localization and detection using
  convolutional networks.
\newblock {\em arXiv preprint arXiv:1312.6229}, 2013.

\bibitem{simonyan2014very}
K.~Simonyan and A.~Zisserman.
\newblock Very deep convolutional networks for large-scale image recognition.
\newblock {\em arXiv preprint arXiv:1409.1556}, 2014.

\bibitem{smith2016deep}
L.~N. Smith and N.~Topin.
\newblock Deep convolutional neural network design patterns.
\newblock {\em arXiv preprint arXiv:1611.00847}, 2016.

\bibitem{song2014learning}
H.~O. Song, R.~Girshick, S.~Jegelka, J.~Mairal, Z.~Harchaoui, and T.~Darrell.
\newblock On learning to localize objects with minimal supervision.
\newblock {\em arXiv preprint arXiv:1403.1024}, 2014.

\bibitem{szegedy2013deep}
C.~Szegedy, A.~Toshev, and D.~Erhan.
\newblock Deep neural networks for object detection.
\newblock In {\em Advances in neural information processing systems}, pages
  2553--2561, 2013.

\bibitem{tang2016large}
Y.~Tang, J.~Wang, B.~Gao, E.~Dellandr{\'e}a, R.~Gaizauskas, and L.~Chen.
\newblock Large scale semi-supervised object detection using visual and
  semantic knowledge transfer.
\newblock In {\em Proceedings of the IEEE Conference on Computer Vision and
  Pattern Recognition}, pages 2119--2128, 2016.

\bibitem{vorontsov2019boosting}
E.~Vorontsov, P.~Molchanov, W.~Byeon, S.~De~Mello, V.~Jampani, M.-Y. Liu,
  S.~Kadoury, and J.~Kautz.
\newblock Boosting segmentation with weak supervision from image-to-image
  translation.
\newblock {\em arXiv preprint arXiv:1904.01636}, 2019.

\bibitem{ward2016general}
L.~Ward, A.~Agrawal, A.~Choudhary, and C.~Wolverton.
\newblock A general-purpose machine learning framework for predicting
  properties of inorganic materials.
\newblock {\em npj Computational Materials}, 2:16028, 2016.

\bibitem{webb2018deep}
S.~Webb.
\newblock Deep learning for biology.
\newblock {\em Nature}, 554(7693), 2018.

\bibitem{xu2015show}
K.~Xu, J.~Ba, R.~Kiros, K.~Cho, A.~Courville, R.~Salakhudinov, R.~Zemel, and
  Y.~Bengio.
\newblock Show, attend and tell: Neural image caption generation with visual
  attention.
\newblock In {\em International conference on machine learning}, pages
  2048--2057, 2015.

\bibitem{yosinski2015understanding}
J.~Yosinski, J.~Clune, A.~Nguyen, T.~Fuchs, and H.~Lipson.
\newblock Understanding neural networks through deep visualization.
\newblock {\em arXiv preprint arXiv:1506.06579}, 2015.

\bibitem{you2016image}
Q.~You, H.~Jin, Z.~Wang, C.~Fang, and J.~Luo.
\newblock Image captioning with semantic attention.
\newblock In {\em Proceedings of the IEEE conference on computer vision and
  pattern recognition}, pages 4651--4659, 2016.

\bibitem{zhang2016yin}
P.~Zhang, Y.~Goyal, D.~Summers-Stay, D.~Batra, and D.~Parikh.
\newblock Yin and yang: Balancing and answering binary visual questions.
\newblock In {\em Proceedings of the IEEE Conference on Computer Vision and
  Pattern Recognition}, pages 5014--5022, 2016.

\end{thebibliography}
}

\end{document}